\pdfoutput=1

\documentclass[11pt]{article}

\usepackage[preprint]{acl}

\usepackage{times}
\usepackage{latexsym}
\usepackage[skins, breakable]{tcolorbox}
\usepackage[T1]{fontenc}

\usepackage[utf8]{inputenc}

\usepackage{microtype}

\usepackage{inconsolata}

\usepackage{graphicx}
\usepackage{listings}
\usepackage{xcolor}

\title{Exploring the Readiness of Prominent Small Language Models for the Democratization of Financial Literacy}

\author{Tagore Rao Kosireddy \and Jeffrey D. Wall \and Evan Lucas \\
  Michigan Technological University / 1400 Townsend Drive \\
  Houghton, Michigan, United States of America \\
  \texttt{{trkosire, jdwall, eglucas} @mtu.edu} \\
}

\begin{document}
\maketitle
\begin{abstract}
The use of \emph{small language models} (SLMs), herein defined as models with less than three billion parameters, is increasing across various domains and applications. Due to their ability to run on more accessible  hardware and preserve user privacy, SLMs possess the potential to democratize access to language models for individuals of different socioeconomic status and with different privacy preferences. This study assesses several state-of-the-art SLMs (e.g., Apple's OpenELM, Microsoft's Phi, Google's Gemma, and the Tinyllama project) for use in the financial domain to support the development of financial literacy LMs. Democratizing access to quality financial information for those who are financially under educated is greatly needed in society, particularly as new financial markets and products emerge and participation in financial markets increases due to ease of access. We are the first to examine the use of open-source SLMs to democratize access to financial question answering capabilities for individuals and students. To this end, we provide an analysis of the memory usage, inference time, similarity comparisons to ground-truth answers, and output readability of prominent SLMs to determine which models are most accessible and capable of supporting access to financial information. We analyze zero-shot and few-shot learning variants of the models. The results suggest that some off-the-shelf SLMs merit further exploration and fine-tuning to prepare them for individual use, while others may have limits to their democratization. Code to replicate our experiments is \href{https://github.com/Tagore-7/Small-Language-Models-for-the-Democratization-of-Financial-Literacy}{shared}\footnote{https://github.com/Tagore-7/Small-Language-Models-for-the-Democratization-of-Financial-Literacy}.
\end{abstract}

\section{Introduction}\label{intro}
 Recent advances in \emph{natural language processing} (NLP) \citep{vaswani2017attention, radford2018improving, devlin2018bert, ethayarajh2019contextual, lewis2019bart, lewis2020pretrained, thoppilan2022lamda, brown2020language, yang2023fingpt} have helped push \emph{artificial intelligence} (AI) as a field into the public sphere of awareness.  A \emph{language model} (LM) can be defined as a statistical model which predicts the conditional probability of a word given some context \citep{bengio2000neural}.

  LMs present an opportunity within the financial sector, where financial knowledge resides primarily with financial professionals. With the advent of new \emph{financial technology} (FinTech) applications, such as online brokerage apps and tax preparation software, the general population has greater access to manage their own finances. FinTech applications, new financial markets (e.g., crypto and NFTs), and social media have facilitated a spike in participation in financial markets \citep{fisch2022gamestop, tinn2021everyone}.    
  
  Increased access to financial markets, the proliferation of new markets, and the constant evolution of financial regulations and codes has brought about unwise use of markets and money. For example, the investment behavior of many individual investors resembles gambling \citep{gao2015individual}, draws on simple heuristics such as mimicking the behavior of social influencers \citep{pedersen2022game}, or even stems from a fear of missing out on social investing trends \citep{pedersen2022game}. Although access to financial markets has improved,  financial literacy is still limited by the scattered and sometimes costly nature of financial data. The current democratization of financial technology is leading to uneducated and potentially risky  behavior by individuals \citep{pedersen2022game, gao2015individual}.

Language models are used in various financial sector activities, such as predictive modeling (stock prediction), portfolio management, financial text mining,  providing financial advice and customer service, picking stocks, and generating meaningful narratives from unstructured financial data  \citep{dredze2016twitter, araci2019finbert, bao2021plato, delucia2022bernice, kim2024financial, zhang2020deep, pagliaro2021investor, gupta2020comprehensive, shahfinaid}. 

Although \emph{large language models} (LLMs) have shown promise in various financial sector activities descried above, these models require access to substantial computing resources or expensive services provided by third parties. For example, Bloomberg GPT is a costly LLM currently utilized by financial professionals that is offered through the Bloomberg terminal \citep{wu2023bloomberggpt}.  These larger models are not particularly accessible to individuals with lower socioeconomic status or low technological capability. Small language models may provide a solution to truly democratize access to language models to support financial literacy in a low-cost and privacy-preserving manner. 

SLMs capable of answering financial questions may also be beneficial for finance, accounting, and economics students. Although some finance programs in the U.S. invest in the Bloomberg terminal, not all universities in the U.S. or globally can afford such tools. SLMs that can run on student computers, lab computers, or even small, cost-effective university servers could provide students with access to SLMs for developing financial literacy and exploring the use of LMs within finance, accounting, and economics. Although it may be difficult or impossible to create an SLM with the same capabilities as an LLM \citep{kaplan2020scaling}, SLMs may still provide value in  democratizing access to information in a low-cost and private manner.

  The term \emph{small language model} (SLM) is not yet well defined in the literature. There is not a consensus on what qualifies as an SLM, yet the topic continues to be of interest to researchers \citep{zhao2023survey, schick2020s, mehta2024openelm}. Herein, we define SLMs as language models with three billion or fewer parameters. This is an arbitrary threshold, but represents an approximate parameter count that can be executable with reasonable inference times on consumer grade technologies, such as personal computers, laptops, and even mobile phones. 
  
  Due to the smaller number of parameters, these models are less resource intensive than their larger counterparts and can be run privately on an individual's computing device, making them an excellent candidate for democratizing LMs for financial literacy. Based on the potential of SLMs for democratization, this study seeks to examine the following research question: are state-of-the-art, off-the-shelf SLMs capable of answering financial questions for individuals with only zero-shot or few-shot learning?

  The rest of this paper is organized as follows. Section \ref{related work} covers related work including a review of existing financial LMs and prominent SLMs. Section \ref{method} describes the selection of SLMs, model parameters, ground-truth data, and comparison criteria for the study. Section \ref{Results} presents the results of the analysis of nine SLM models based on criteria related to memory usage, inference time, similarity measures, and a readability test. Section \ref{cons} summarizes our work, identifies which SLM performed the best on the selected criteria, and makes suggestions for future efforts. In addition to the core content of our paper, a section for Limitations and an Ethics statement are included alongside Acknowledgements at the end of the paper.

\section{Related Work} \label{related work}
Our work explores the application of SLMs in the support of financial literacy, which branches across research on financial language models and on SLMs.  

\subsection{Existing Financial Language Models}
Bloomberg recently introduced a 50 billion parameter transformer model trained on 363 billion tokens from the company's finance-specific  text resources and 345 billion tokens from general purpose text datasets \citep{wu2023bloomberggpt}. This model, BloombergGPT, is a commercially available model that is available through the Bloomberg terminal. Access to the model, which includes high pricing and licensing agreements, limits the democratization of its use.

In an effort to further democratize LLMs for finance, a group of researchers developed the open source FinGPT model \citep{yang2023fingpt}. FinGPT was designed as a full-stack application including a layer for the open source financial data sources, a data engineering layer, a layer for retrieval augmented generation language models, and an application layer with simple web-based demos. The primary model explored in the seminal FinGPT paper was based on a Llama-7B model. The model was trained on financial news, corporate financial statements, and other sources \citep{yang2023fingpt}. The various fine-tuned FinGPT LLMs are available through the FinGPT GitHub page \citep{githubGitHubAI4FinanceFoundationFinGPT}. However, FinGPT models are only trained for predictive and classification tasks, not for text generation tasks that would support financial literacy.

Other similar models include InvestLM and FinMA. InvestLM is a fine-tuned LLaMA-65 billion parameter model tuned on a variety of financial data sources \citep{yang2023investlm}. Similarly, FinMA is a series of fine-tuned LLama model with 7 and 13 billion parameters that were also trained on a variety of financial data sources \citep{xie2023pixiu}. Some of these models provide text generation capabilities, but the original datasets are not open-sourced, likely because some of the textbooks and other sources area copyrighted. These lack of open datasets limit the ability to further refine and develop these models. 

Other recent advances in financial LLMs, such as FinMem, provide memory layers that allow the model to more accurately draw insight from real time financial data, as financial investment is dynamic in nature \citep{yu2024finmem}. 

Although many studies \cite{xing2018natural, araci2019finbert, liu2022finrl, shah2022flue, yang2023data, guo2023chatgpt, zhang2023xuanyuan, li2023chatgpt, yang2023fingpt, wu2023bloomberggpt, li2023large, xie2023pixiu, huang2023finbert, lee2024survey, kim2024financial} examine LLMs in the finance domain, to the best of our knowledge, we are the first to examine the application of SLMs to enhance the financial literacy of individuals and students using solely open-source resources. 

Open models like FinGPT, InvestLM, and FinMA offer a positive step toward democratizing financial LLMs. However, many of the current models may be too large for most individuals and students to use on their limited computing devices, don't offer text generation capabilities, or are based on copyrighted datasets that limit further development. We call for other researchers to explore financial applications of open-source SLMs and finance question-answering datasets to further democratize access to and refinement of finance language models.

\subsection{State-of-the-art small language models} 
At the time of writing, we identified several promising SLMs to include in our work, which have recently been developed by Google, Microsoft, and Apple, and models built from Meta's open sourced Llama models.
All of these models have achieved state-of-the-art results for SLMs and are well-supported by organizations or communities. 
In this work, we focus solely on open source SLMs, as our larger motive is the democratization of language models. 

The Gemma language models were developed by the Google Gemma Team and Google DeepMind based on Google's Gemini LLMs \citep{team2024gemma}. The Gemma models were designed to be much smaller than the Gemini models, with two models consisting of a seven billion parameter model and a smaller two billion parameter model. The two billion parameter model was designed for consumer-grade computing devices, meeting our criteria. The Gemma models are designed on sequence models and transformers. The two billion parameter model consists of 18 layers and 8 heads of size 256. The model was trained on three trillion tokens from web documents, and mathematical and code resources. The Gemma models perform well on common benchmarks like \emph{Multi-task Language Understanding} (MMLU), as compared to similarly sized models (i.e., 2-13 billion parameters) \citep{team2024gemma}.

The Phi family of language models were developed by Microsoft. Microsoft introduced Phi-3 in 2024, which extended earlier work on the Phi-1 and Phi-2 models \citep{abdin2024phi}. Phi-3, along with the rest of the Phi family, is based on the transformer architecture and was built to be compatible with Llama models, such as using the same tokenizer. The introduction of the Phi-3 model consisted of 32 layers and 32 heads and more than three billion parameters, which is larger than Phi-1 and Phi-2. Microsoft has since included multiple model sizes. Phi-3 was also trained on a larger dataset (more than three trillion tokens) than its predecessors, which included both web data and synthetic data. Like the Gemma models, the Phi models have reached state-of-the-art results on common benchmarks.

The OpenELM models were developed by Apple. Apple introduced the OpenELM three billion parameter model in 2024, with smaller models with as few as 270 million parameters . The OpenELM models also rely on the transformer architecture. Similar to the Phi models, the OpenELM models utilize the same tokenizer as the Llama models for compatibility. Unlike other models, OpenELM utilizes a variable number of heads for each layer. The models were trained on a variety of common data sets with more than one trillion tokens. Like the other models we discuss, the OpenELM models offer state-of-the-art results on common benchmarks \citep{mehta2024openelm}.

TinyLlama is an open source model inspired by Meta's Llama models \citep{zhang2024tinyllama}. Although TinyLlama was not developed by Meta directly, its foundation on the Llama models grants it a strong support community and compatibility with many other models. Like other Llama models, TinyLlama relies on the transformer architecture. The model consists of 22 layers and 32 heads. The model was trained on three trillion tokens from two primary sources that contained natural language and code. TinyLlama also boasts state-of-the-art results on common benchmarks \citep{zhang2024tinyllama}.

Fine-tuned SLMs have performed reasonably well on a variety of tasks across many domains, such as meeting summarization \citep{fu2024tiny}, hate speech detection \citep{sen2024hatetinyllm}, and radiology question answering \citep{ranjit2024rad}. The SLMs in our study represent potential candidates for developing financial literacy SLMs. We now explain our study design used to evaluate the potential of these models.

\section{Method}\label{method}
 To evaluate whether state-of-the-art SLMs are prepared to answer financial questions, we: 1) identified a set of state-of-the-art open source SLMs, 2) selected important criteria to evaluate the model output to ensure the outputs were accessible to individuals from different socioeconomic groups and education levels, 3) identified a set of open-source question/answer pairs to evaluate the models, and 4) conducted a study to determine how well each model performed on the selected criteria.

\subsection{Model details}
 To initiate our study of SLMs for financial literacy, we developed a list of prominent, state-of-the-art small language models with three billion parameters or less, namely Apple's OpenELM(270M, 450M, 1.1B, 3B) \citep{mehta2024openelm}, Microsoft's Phi(1B, 1.5B, 2B) \citep{gunasekar2023textbooks, li2023textbooks, javaheripi2023phi}, Google's Gemma \citep{team2024gemma}, and the TinyLlama \citep{zhang2024tinyllama} models. We downloaded all models and the dataset from HuggingFace.  
 
 We selected these models as they are all small (<3B) open-source models created by corporations or communities that are likely to support their further development. For example, the Gemma, OpenELM, and Phi models are supported by large technology corporations. Similarly, TinyLlama is based on Meta's open source Llama models, which has a large support community.

 We also tried to limit the selected models to the pre-trained model versions that didn't have additional tuning with chat or instruction training data. We did this to ensure the models were as comparable as possible.  
 For example, we used the model resulting from the last training step of the TinyLlama model rather than the chat version. Similarly, we only included Microsoft's Phi-1B, Phi-1.5B, and Phi-2B models. We found no generalized LM version of Phi-3.

\subsection{Model evaluation dataset}
 We also identified a series of financial questions, with their answers, that an individual might have to improve their financial literacy. We used the question/answer pairs to verify the quality of the SLM responses. The open-source "FinGPT/fingpt-fiqa\_qa" dataset on HuggingFace from the FinGPT project \cite{yang2023fingpt} was selected for this purpose. The FinGPT model and datasets have been vetted in multiple studies \citep{yang2023fingpt, zhang2023instructfingpt, zhang2023fingptrag, wang2023fingptbenchmark, 2023finnlp}. The dataset contains questions asked by novice users on financial forums along with reasonable responses provided by forum participants. The dataset also contained system prompts for prompt engineering that we used in each prompt. This dataset is the only open-source and available question/answer financial dataset in existence. Other studies of open-source financial models have used custom  datasets for training, such as for the FinMA model. This data is not openly available, likely due to the use of copyrighted materials.
 
 For the purposes of this study, we treated the answers to each question in the dataset as the ground truth for comparison with the responses from the SLMs. After removing duplicate questions, the dataset consisted of 6105 question/answer pairs. 
 
 The usage of multiple and varied input prompts is important for extracting useful information from a language model \citep{liu2022design, zhou2022large, white2023prompt}.  As such, we randomly sampled 100 question/answer pairs from the dataset with random seed 7. These 100 question/answer pairs were used for the remainder of the study. 

\subsection{Model parameters}
 To ensure that the responses from each model were comparable, we used the same generation parameters for all of the SLMs (max\_new\_tokens=250, top\_k = 30, top\_p = 0.8, no\_repeat\_ngram\_size=5). The max\_new\_tokens was determined by averaging token length of the ground truth answers from the dataset (252.26 tokens), which we rounded to  250. We used the Llama-2-7b-chat-hf tokenizer to calculate the average token length. Prior to the study, the model parameter values were identified by evaluating the quality of the outputs generated by the models through simple trial and error experimentation by one of the authors.
 
\subsection{Computing resources}
All of the analysis was conducted in Google Colab in Python using the following computational resources: GPU 15(GiB) and RAM 12.7 (GB). Due to limited computational resources, models with more than 1.5 billion parameters were loaded in half-precision using bf16 \citep{kalamkar2019study} to mimic consumer-grade technology limitations. We chose to use Colab over our more powerful research computers or computing cluster because the limited computing resources of Google Colab's free tier better represents the consumer electronics used by retail investors (and in fact is an easily available resource an individual could use!). The results of this analysis are limited to the use of the SLMs with the minimum computational resources  shown in Table \ref{table:modelmemoryrequirements}.

\subsection{Model comparison criteria}
 To compare the responses from each model, each of the 100 sample questions was provided to each of the nine models in both a zero-shot and few-shot in-context learning approach.
 
 The outputs were evaluated against the ground truth answers from the FinGPT dataset by calculating the similarity between the outputs and the ground truth answers, as presented in Table \ref{table:outputsimilarityscores}. 
 
 To compare the model outputs with the ground truth answers, several similarity comparison metrics were calculated. First,  we calculated the Semantic Textual Similarity (STS) using Cross-Encoder, which achieves better performance than Bi-Encoder \citep{reimers2019sentence, risch2021semantic}. STS utilizes sentence transformer models to convert text into vectors (embeddings) that capture semantic information about the text  \citep{reimers-2019-sentence-bert}, providing a similarity score between 0 and 1. Second, we calculated several ROUGE metrics, which are commonly used to evaluate the degree of overlap in words, bi-grams, or common substrings between a candidate and reference sentence (or sentences) \citep{lin2004ROUGE}. ROUGE metrics range between 0 and 1, with higher scores indicating higher similarity between the automatically produced summary and the reference. The ROUGE scores do not take semantic meaning into account and have been criticized for this shortcoming \cite{akter2022revisiting}. Third, we calculated the BERTScore for the models outputs, which measures the similarity between a candidate sentence or sentences and a reference sentence or sentences using contextual embeddings \citep{zhang2019bertscore}, resulting in scores between -1 and 1. The BERTScore is less sensitive to smaller errors, especially if the candidate text is lexically or stylistically similar to the reference text \citep{hanna2021fine}.
 
 To better understand the computing requirements for each model, we calculated GPU and memory usage and inference times. We monitored GPU consumption with the NVIDIA CUDA library and Google Colab's built-in resource graphs. Memory calculations during the model loading process were assessed with the psutil Python library. We loaded each model ten times, as memory consumption differed slightly each time. We started a new session and new fresh runtime each time we loaded the model. We then calculated the average values as presented in Table \ref{table:modelmemoryrequirements}. The inference time was calculated by taking the average inference time of the 100 sample questions. These sample questions were the same question/answer pairs selected from the dataset for the remainder of the study.

We also included an analysis of the readability of the model outputs. Since our goal is to assess the use of SLMs for financial literacy from a democratization lens, producing content that is readable by individuals with lower reading levels is important. Models that produce outputs that require a college reading level may not be ideal for the democratization of models. Readability was calculated using the widely used Flesch–Kincaid readability test \citep{flesch2007flesch}. The test examines the lexical complexity of text. The resulting values range from 0 to 100, with values near 0 representing complex and difficult to read text and scores near 100 representing easy to read text. To assess readability, we combined the responses for each model to produce a single readability score for each model. Combining each models' outputs was necessary due to the input length requirements of the Flesch–Kincaid readability test. The readability scores are presented in Table \ref{table:modelreadabilityscores}.

\begin{table*}[ht!]
\centering
\caption{Model Memory Requirements and Inference Time}\label{table:modelmemoryrequirements}
\begin{tabular}{|l|c|c|c|c|}
\cline{1-5}
\textbf{Model}    & \textbf{GPU(GiB)}   &  \textbf{RAM(MB)} & \textbf{Avg. Inf Time(sec)}& \textbf{ Precision}\\ \cline{1-5}
(1) Apple-OpenELM-270M   &   2.2     &    642.2977      & 5.64 &  full \\ \cline{1-5}
(2) Apple-OpenELM-450M  &   3.7   &   588.7348    &  7.32 &  full \\ \cline{1-5}
(3) Apple-OpenELM-1.1B  &    8.2  &     765.3945   &  9.89 &  full \\ \cline{1-5}
(4) Apple-OpenELM-3B    &  13.6    &  473.3031      &  14.60 &  half \\ \cline{1-5}
(5) Microsoft-phi-1B     &    8.2  &   759.8051    & 7.28 &  full \\ \cline{1-5}
(6) Microsoft-phi-1.5B        &  8.2    &  670.2625      &  7.30 &  full \\ \cline{1-5}
(7) Microsoft-Phi-2B          &   10.3   &   410.8238     &  7.07 &  half \\ \cline{1-5}
(8) Google-gemma-2B           &   9.5   &   792.9766     &  6.68 &  half \\ \cline{1-5}
(9) TinyLlama-1.1B   &  8.3    &  721.0668      &  5.65 &  full \\ \cline{1-5}
\end{tabular}
\end{table*}

\begin{table*}[ht!]
\centering
\caption{  Similarity Scores Between Output and Ground Truth (Mean Zero-Shot * Mean Few-Shot)}\label{table:outputsimilarityscores}
\begin{tabular}{|l|c|c|c|c|c|}
\hline
\textbf{Model} & \textbf{STS} & \textbf{ROUGE-1} & \textbf{ROUGE-2} & \textbf{ROUGE-L} & \textbf{BERTScore}  \\ \hline
(1) & \(0.5142  *  0.4997\) & \(0.2497  * 0.2533\) & \(0.0362 * 0.0379\) & \(\textbf{0.1392} * 0.1316\) & \(0.8165 * 0.8220\)  \\ \hline
(2) & \(0.5214 * 0.5113\) & \(0.2303 * 0.2487\) & \(0.0285 * 0.0359\) & \(0.1305 * 0.1293\) & \(0.8140 * 0.8215\) \\ \hline
(3) & \(0.5010 * 0.5228\) & \(0.2533 * 0.2579\) & \(0.0373 * 0.0401\) & \(0.1364 * 0.1327\) & \(0.8170 * 0.8246\) \\ \hline
(4) & \(0.4970 * 0.5048\) & \(0.2469 * 0.2445\) & \(0.0363 * 0.0372\) & \(0.1317 * 0.1283\) & \(0.8165 * 0.7991\) \\ \hline
(5) & \(0.5094 * 0.4876\) & \(0.1699 * 0.2251\) & \(0.0125 * 0.0280\) & \(0.0958 * 0.1181\) & \(0.7675 *  0.7966\) \\ \hline
(6) & \(0.4838 * \textbf{0.5403}\) & \(0.2164 * 0.2515\) & \(0.0244 * 0.0364\) & \(0.1131 * 0.1279\) & \(0.8075 * 0.8222\) \\ \hline
(7) & \(0.5222 * 0.5370\) & \(0.2390 * 0.2485\) & \(0.0402 * \textbf{0.0429}\) & \(0.1279 * 0.1294\) & \(0.8091 * 0.8040\) \\ \hline
(8) & \(0.4797 * 0.5299\) & \(0.2013 * \textbf{0.2683}\) & \(0.0250 * 0.0428\) & \(0.1168 * 0.1367\) & \(0.8106 * \textbf{0.8260}\) \\ \hline
(9) & \(0.4842 * 0.4987\) & \(0.1970 * 0.2626\) & \(0.0282 * 0.0390\) & \(0.1082 * 0.1320\) & \(0.8136 * 0.8229\) \\ \hline
\end{tabular}
\end{table*}

\begin{table*}[ht!]
\centering
\caption{Readability Scores (Zero-shot score * Few-shot Score * Change)}\label{table:modelreadabilityscores}
\begin{tabular}{|l|l|}
\hline
\textbf{Model}    & \textbf{Readability Score} \\ \hline
(1) Apple-OpenELM-270M & 74.54 * \textbf{70.45} * -4.09 \\
\hline
(2) Apple-OpenELM-450M & 73.13 * 69.20 * -3.93 \\ \hline
(3) Apple-OpenELM-1.1B & 72.78 * 68.23 * -4.55 \\ \hline
(4) Apple-OpenELM-3B & 74.47 * 68.53 * -5.94 \\ \hline
(5) Microsoft-phi-1B & 54.64 * 59.30 * +4.66\\ \hline
(6) Microsoft-phi-1.5B & 54.90 * 57.70 * +2.80\\ \hline
(7) Microsoft-Phi-2B & 60.03 * 58.95 * -1.08\\ \hline
(8) Google-gemma-2B & \textbf{78.44} * 65.60 * -12.84 \\ \hline
(9) TinyLlama-1.1B-intermediate-step-1431k-3T & 74.49 * 68.99 * -05.50 \\ \hline
\end{tabular}
\end{table*}

\subsection{Zero- and few-shot learning}
For this study, we ran each of the SLMs with a zero-shot approach, including only a simple instruction prompt included with the dataset, and again with in-context few-shot learning without the instruction prompt. We used five few-shot learning examples designed by the business researcher on the research team. After crafting the examples, they were passed through ChatGPT-4o with a request to make the writing accessible to individuals with a high-school education. We did this to create a reasonably readable set of few-shot learning examples. The Flesch-Kincaid readability score for the examples was 60.27, which approximates a high-school reading level.  

\section{Results}\label{Results}
As outlined in the following sections, the models possess different strengths and limitations that affect their ability to be utilized by individuals on consumer-grade electronics to develop financial literacy. We assess each model on a variety of aspects, including memory usage, inference time, capability of answering financial questions, and the readability of model outputs.

\subsection{Memory use and inference time}
The consumption of Graphical Processing Units (GPUs) and Random Access Memory (RAM) is denoted in \emph{gigabytes} (GiB) and \emph{megabytes} (MB) respectively for all the models in Table \ref{table:modelmemoryrequirements} along with  average inference time (seconds).

The GPU usage (GiB) ranged from 2.3 for the OpenELM-270M model to 13.6 GiB for the OpenELM-3B model. Clearly, the OpenELM-270M and -450M models provide the best support for low-grade consumer electronics with GiB requirements below 4 GiB. However, the models that are lower than or near 8 GiB are reasonable for some consumer-grade laptops and personal computers. The larger models (i.e., OpenELM-3B, gemma-2B, and Phi-2) with requirements well above 8 GiB may only be ideal for those investors with the means to purchase adequate GPUs. For all models, RAM usage did not exceed numbers that would be considered excessive for consumer-grade electronics. 

Average inference times ranged from 5.65 seconds for the OpenELM-270M model to 14.60 seconds for the OpenELM-3B model at half-precision. Given that the Google Colab GPU may be slightly more powerful than many retail investor's GPUs, the models with inference times much above 7 seconds could be excessive for queries made on some consumer-grade devices. Thus, based on inference speeds alone, OpenELM-1.1B and -3B models may not be the most appropriate for retail investors with low-grade GPUs.

\subsection{Financial question answering similarity comparisons}
All of the similarity scores (i.e., semantic textual similarity (STS), ROUGE Scores, and BERTScore) are presented in Table \ref{table:outputsimilarityscores} and in extended form in Table \ref{table:extendedoutputsimilarityscores}. 

 The Semantic Textual Similarity (STS) showed medium and low standard deviations for all models. The highest mean STS is the Phi-1.5B few-shot model at 0.5403, while the lowest mean STS is the Gemma-2B zero-shot model. The top four performing models were all few-shot learning models, namely the Phi-1.5B few-shot (0.5403), Phi-2B few-shot (0.5370), Gemma-2B few-shot (0.5299), and OpenELM-1.1B few-shot (0.5228) models. 

The highest ROUGE-1 mean score is 0.2683 for the Gemma-2B few-shot model, and the lowest is 0.1699 for the Phi-1B zero-shot model. The top four performing models are also all few-shot learning models, namely the Gemma-2B few-shot (0.2683), TinyLlama-1.1B few-shot (0.2626), OpenELM-1.1B few-shot (0.2579), and OpenELM-270M few-shot (0.2533) models.

The highest ROUGE-2 mean score the Phi-2B few-shot model at 0.0429, and the lowest is Phi-1B zero-shot at 0.0125. Three of the top four performing models were few-shot models, including the Phi-2B few-shot (0.0429), Gemma-2B few-shot (0.0428), Phi-2B zero-shot (0.0402), and OpenELM-1.1B few-shot (0.0401). 

The highest ROUGE-L mean score is the OpenELM-270M zero-shot model at 0.1392, while the lowest is the Phi-1B zero-shot model at 0.0958. The top four performing models are half zero-shot and half few-shot models, namely OpenELM-270M zero-shot (0.1392), Gemma-2B few-shot (0.1367), OpenELM-1.1B zero-shot (0.1364), and OpenELM-1.1B few-shot (0.1327) models.

The highest BERTScore F1 mean score is the Gemma-2B few-shot model at 0.8260, and the lowest is Phi-1B zero-shot model at 0.7675. The top four performing models were all few-shot models, including the Gemma-2B few-shot (0.8260), OpenELM-1.1B few-shot (0.8246), TinyLlama-1.1B few-shot (0.8229), and Phi-1.5B few-shot (0.8222) models.

The results show that few-shot learning generally improved the similarity comparisons for most models. Six of the nine models had higher STS for the few-shot variants; eight of the models had higher ROUGE-1 scores for the few-shot variants; all of the models had higher ROUGE-2 scores for the few-shot variants; five of the models had higher ROUGE-L scores for the few-shot variants; and seven of the models had higher BERTScores for the few-shot variants. The OpenELM models exhibited the greatest number of comparison metric scores decreased after few-shot learning. The Phi-1.5B, Gemma-2B, and TinyLlama1.1B models showed improvement across all comparison metrics after using few-shot learning.

\subsection{Readability scores}
The Gemma-2B zero-shot model provides the most readable output with a score of 78.44, which approximates a middle school reading level. The least readable model responses were provided by Microsoft's Phi-1B model with a score of 54.64, which approaches college level reading levels. The top four performing models in terms of readability are the Gemma-2B zero-shot (78.44), OpenELM-270M zero-shot (74.54), TinyLlama-1.1B zero-shot (74.49), and OpenELM-3B zero-shot (74.47) models. 

All of the OpenELM models, the Gemma-2B model, and the TinyLlama-1.1B model showed a decrease in readability after few-shot learning, ranging from a decrease in 3.93 for the OpenELM-450M model to 12.84 for the Gemma-2B model. Two of the Phi models saw an increase in readability; Phi-1B with a 4.66 increase and Phi-1.5B with a 2.8 increase. The Phi-2B model saw a modest decrease of 1.08 after few-shot learning. The changes in readability are likely due to the readability of the few-shot learning examples, which had a score of 60.27. Overall, the few-shot learning examples pushed all of the models closer to a high-school reading level (~60.00), retaining or improving accessibility of the models.

\section{Conclusions}\label{cons}
The results of the study suggest that certain SLMs are better candidate models for future fine-tuning and development to support the democratization of financial literacy LMs than others. See Appendix B for further comparisons.

Several of Apple's OpenELM models show great promise for future study of the democratization of financial literacy LMs. The low memory requirements and inference times, and higher readability scores exhibited by these models make them ideal for democratization. Of these models, the OpenELM-270M model provides the greatest accessibility in terms of GPU requirements, inference times, and readable outputs. The OpenELM-270M zero-shot and few-shot models also scored in the top four performers on at least one of the similarity score metrics, which larger counterparts (e.g., the OpenELM-450M and OpenELM-3B models) did not achieve. The OpenELM-1.1B few-shot model showed very promising scores across the similarity comparison metrics, scoring in the top four models for all of the similarity metrics. However, the higher GPU requirements and inference time limits its accessibility and usefulness. 

The Microsoft Phi models also show some promise for future study in this domain, but likely only for college educated individuals. Although the Phi-1B model exhibited some of the worst comparability metric scores of all of the models and the lowest zero-shot readability score, the Phi-1.5B and Phi-2B few-shot models both appeared in the the top four performers on two  similarity comparison metrics. Of course, these models also exhibited low readability scores, which changed only slightly after few-shot learning. The other major limiting factor of the Phi models is their higher GPU requirements. Although the Phi models show promise for future financial literacy models, they may be better suited for college students or graduates than by those with a lower reading level.

Google's Gemma model also shows promise for moderately powerful consumer-grade technology. It had faster inference times than most of the models. The zero-shot model also had the best readability score, making it the most accessible in terms of reading level. The Gemma-2B model also exhibited some of the best similarity comparison scores, scoring in the top four for all metrics, though only after few-shot learning. This model could be ideal for individuals with access to respectable consumer-grade computing devices. Like Gemma-2B, the TinyLlama-1.1B model had good inference speeds and reasonable readability scores. The few-shot model also appeared in the top four performers on similarity comparison metrics twice. Also like Gemma-2B, the TinyLlama-1.1B model suffers from higher GPU memory requirements. 

Overall, this study suggests that Apple's new OpenELM-270M model deserves further research attention from the lens of democratizing language models for the greatest number of individuals. This model has a small memory profile, fast inference times, produces reasonable results as compared to ground-truth finance responses, and produces reasonable readability scores. However, in some cases, such as for individuals with more powerful computing devices,   the OpenELM-1.1B and Gemma-2B models share similar GPU requirements, inference times, and high readability and similarity comparison scores. The Phi-1.5B and Phi-2B models may also be useful for college educated individuals. 

Society is on the cusp of democratizing financial investment information, which has long been limited to the financial elite. We encourage researchers to continue to explore SLMs and further fine-tune models like OpenELM to support the development of financial literacy for the greatest population possible. We provide a starting point for future research by identifying the most promising models that meet criteria for truly democratized models. 

\section*{Limitations}\label{limitations}
The memory calculation during the model loading process was calculated with the Psutil Python library. We noticed that each time the code was executed, it yielded slightly different memory consumption values. It's important to note that these calculations are only approximations. Additionally, we performed the calculations only for loading the model, not at inference time. 

Our purpose in conducting this study was to assess the feasibility of using SLMs for financial literacy question answering. Existing research is clear that SLMs do not perform at the level of LLMs. As such, we did not compare our results to LLMs like ChatGPT-4o, Claudia, Llama, and others. We acknowledge that these large models perform better than SLMs, but they fail to meet the requirements for democratization laid out in this study.

Given the limited availability of open-source question answering datasets (only one such dataset exists), we did not assess the factuality of responses. The only existing dataset is based on social media opinion. Better open-source financial question answering datasets are required to fully fine-tune and assess the factuality of future models. Existing open-source datasets are not ready to support financial question answering models given the legal and ethical pitfalls in offering sound financial advice. Future research will need to establish higher quality financial question answering datasets, develop knowledge graphs and RAG pipeline to produce more consumer-ready models. This study was designed to test whether researchers should invest in such efforts with existing SLMs models. We showed that models can be improved for financial literacy with even just five higher-quality few-shot learning examples. Further improvements are more than likely if future research seeks to develop a high-quality, open-source financial question answering dataset.

We also did not include human review of the model responses due to time and budgetary limitations. However, we utilized similarity scores that have shown strong correlations with human judgement, although they do not capture important concerns such as toxicity of the answer. A full human review wasn't warranted given the early stage of research in this area. Better datasets need to be created first.

Further, we did not include the Microsoft Phi-3 model, as we were not able to find an untuned version of the model. Using the Phi-3 instruction tuned model could have granted the model an unfair advantage or disadvantage compared to the other models. 

As with other language models, SLMs are subject to special security concerns and hallucinations. We did not explore issues with hallucinations, nor with security issues that could arise with SLMs on consumer devices. Future research should explore the occurrence of financial hallucinations in SLMs, as security and accuracy are as important to the democratization of language models as accessibility and readability. However, such efforts will require the creation of better open-source datasets to properly fine-tune models and develop RAG pipelines. 

\section*{Ethics Statement} \label{sec:ethics}
During the course of this work, we were careful in our selection of data. We selected our data from previously peer-reviewed sources, namely from the FinGPT open sourced data sets. FinGPT and its data sets have been vetted in multiple peer-reviewed publications. 

We also did our best to be as inclusive as possible in our definition of democratization and the selected metrics. For example, we included measures of readability to account for individuals who are systematically limited in their attainment of higher education. Similarly, we were careful in our conclusions to account for socioeconomic status and the availability of different levels of consumer-grade computing devices. We tried to outline which models, given their memory requirements, are best fitted for the broadest user base and which would require access to more expensive consumer devices.

Further, we have outlined some of the current limitations in the field related to the development of finance LMs, such as concerns with the factuality of existing datasets. Although some SLMs deserve further development and testing within the finance domain, core open-source data infrastructure is needed to support such efforts. No one should take the findings of this study to suggest that even few-shot learning is enough to produce good SLMs ready for consumer use. 

\section*{Acknowledgments}\label{sec:acknowledgements}
This study was made possible with support from Michigan Technological University's Institute of Computing and Cybersystems and the College of Business.

\bibliography{acl_latex}

\appendix

\section{Appendix A: Extended Similarity Comparison Table}
\label{sec:appendixExtendedData}
Table \ref{table:extendedoutputsimilarityscores} provides an extended view of the similarity comparison results with separate rows for the zero-show and few-shot model variants. The extended table also includes mean and standard devaiation for each model.

\begin{table*}[ht!]
\centering
\caption{  Similarity Scores between output and ground truth with Avg. \(Mean \pm Std\) for Zero- and Few-shot outputs}\label{table:extendedoutputsimilarityscores}
\begin{tabular}{|l|c|c|c|c|c|}
\hline
\textbf{Model} & \textbf{STS} & \textbf{ROUGE-1} & \textbf{ROUGE-2} & \textbf{ROUGE-L} & \textbf{BERTScore}  \\ \hline
(1) Zero & \(0.5142 \pm 0.0923\) & \(0.2497 \pm 0.1144\) & \(0.0362 \pm 0.0282\) & \(\textbf{0.1392} \pm \textbf{0.0943}\) & \(0.8165 \pm 0.0231\)  \\ \hline
(1) Few & \(0.4997 \pm 0.0718\) & \(0.2533 \pm 0.0943\) & \(0.0379 \pm 0.0275\) & \(0.1316 \pm 0.0415\) & \(0.8220 \pm 0.0158\) \\ \hline
(2) Zero & \(0.5214 \pm 0.0865\) & \(0.2303 \pm 0.1163\) & \(0.0285 \pm 0.0236\) & \(0.1305 \pm 0.0956\) & \(0.8140 \pm 0.0262\) \\ \hline
(2) Few & \(0.5113\pm 0.0807\) & \(0.2487 \pm 0.0911\) & \(0.0359 \pm 0.0266\) & \(0.1293 \pm 0.0390\)& \( 0.8215\pm 0.0171\) \\ \hline
(3) Zero & \(0.5010 \pm 0.0921\) & \(0.2533 \pm 0.1226\) & \(0.0373 \pm 0.0289\) & \(0.1364 \pm 0.0960\) & \(0.8170 \pm 0.0262\) \\ \hline
(3) Few & \( 0.5228 \pm 0.0699\) & \(0.2579 \pm 0.0879\) & \(0.0401 \pm 0.0276\) & \(0.1327 \pm 0.0380\) & \(0.8246 \pm 0.0166\) \\ \hline
(4) Zero& \(0.4970 \pm 0.1012\) & \(0.2469 \pm 0.0912\) & \(0.0363 \pm 0.0279\) & \(0.1317 \pm 0.0416\) & \(0.8165 \pm 0.0208\) \\ \hline
(4) Few & \(0.5048 \pm 0.0796\) & \(0.2445 \pm 0.0998\) & \(0.0372 \pm 0.0289\) & \(0.1283 \pm 0.0479\) & \(0.7991 \pm 0.1416\) \\ \hline
(5) Zero& \(0.5094 \pm 0.0623\) & \(0.1699 \pm 0.0592\) & \(0.0125 \pm 0.0127\) & \(0.0958 \pm 0.0304\) & \(0.7675 \pm 0.0163\) \\ \hline
(5) Few& \(0.4876 \pm 0.0566\)& \(0.2251 \pm 0.0790\) & \(0.0280 \pm 0.0215\) & \(0.1181 \pm 0.0322\) & \( 0.7966\pm 0.0207\) \\ \hline
(6) Zero & \(0.4838 \pm 0.1041\) & \(0.2164 \pm 0.0822\) & \(0.0244 \pm 0.0196\) & \(0.1131 \pm 0.0379\) & \(0.8075 \pm 0.0231\) \\ \hline
(6) Few & \( \textbf{0.5403} \pm \textbf{0.0711}\) & \(0.2515 \pm 0.0892\) & \(0.0364 \pm 0.0253\) & \(0.1279 \pm 0.0378\)& \(0.8222 \pm 0.0156\) \\ \hline
(7) Zero& \(0.5222 \pm 0.0830\) & \(0.2390 \pm 0.1014\) & \(0.0402 \pm 0.0312\) & \(0.1279 \pm 0.0491\) & \(0.8091 \pm 0.1165\) \\ \hline
(7) Few& \(0.5370 \pm 0.0726\)& \(0.2485 \pm 0.1106\) & \(\textbf{0.0429} \pm \textbf{0.0320}\) & \(0.1294 \pm 0.0500\) & \(0.8040 \pm 0.1423\) \\ \hline
(8) Zero & \(0.4797 \pm 0.0936\) & \(0.2013 \pm 0.0766\) & \(0.0250 \pm 0.0218\) & \(0.1168 \pm 0.0359\) & \(0.8106 \pm 0.0226\) \\ \hline
(8) Few & \(0.5299 \pm 0.0661\) & \(\textbf{0.2683} \pm \textbf{0.0885}\) & \(0.0428 \pm 0.0296\) & \(0.1367 \pm 0.0384\) & \(\textbf{0.8260} \pm \textbf{0.0158}\) \\ \hline
(9) Zero & \(0.4842 \pm 0.0878\) & \(0.1970 \pm 0.1098\) & \(0.0282 \pm 0.0257\) & \(0.1082 \pm 0.0506\) & \(0.8136 \pm 0.0175\) \\ \hline
(9) Few & \(0.4987 \pm 0.0760\) & \( 0.2626 \pm  0.0927\) & \(0.0390 \pm 0.0263\) & \(0.1320 \pm 0.0384\) & \(0.8229 \pm 0.0148\) \\ \hline
\end{tabular}
\end{table*}

\section{Appendix B: Performance Tradeoff}
\label{sec:appendixgraph}
In addition to the tabular data presented in the main body of the paper, this appendix presents scatter plots that compare the SLMs based on some criteria combinations.

\begin{figure}[ht!]
\centering
\includegraphics[width=6.5cm]{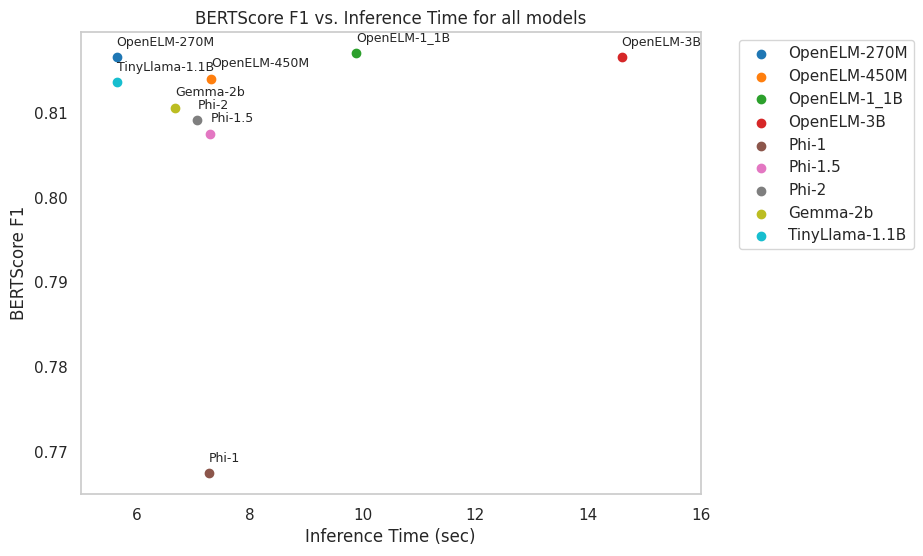}
\caption{BERTScore vs. Inference Time} \label{fig:bertvsInf}
\end{figure}

Figure \ref{fig:bertvsInf} shows a grouping of similar models with higher BERTScores and lower inference times.

\begin{figure}[ht!]
\centering
\includegraphics[width=6.5cm]{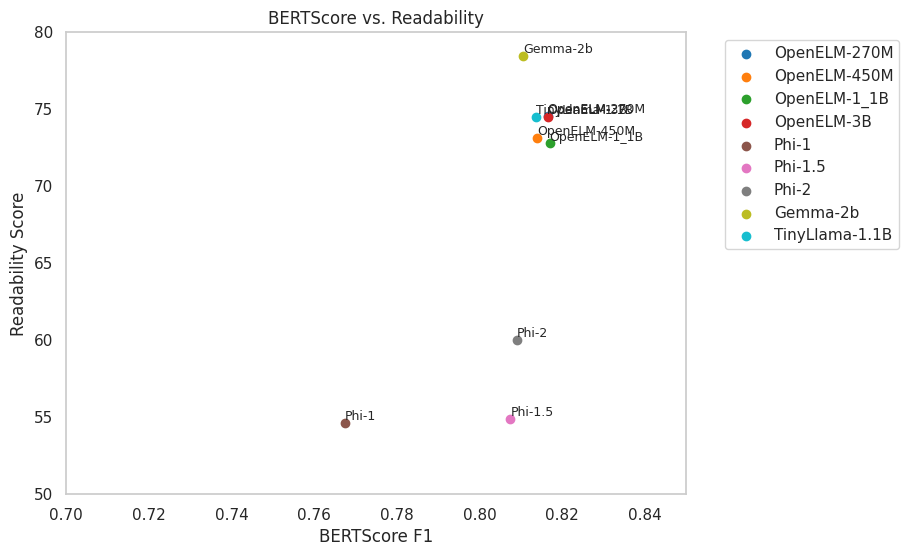}
\caption{ BertScore vs. Readability for zero-shot prompting} \label{fig:bertvsRead}
\end{figure}

Figure \ref{fig:bertvsRead} shows a grouping of models with high readability scores and higher BERTScores, with the Phi models lower on the readability scale. With a few models is less desirable positions.

\begin{figure}[ht!]
\centering
\includegraphics[width=6.5cm]{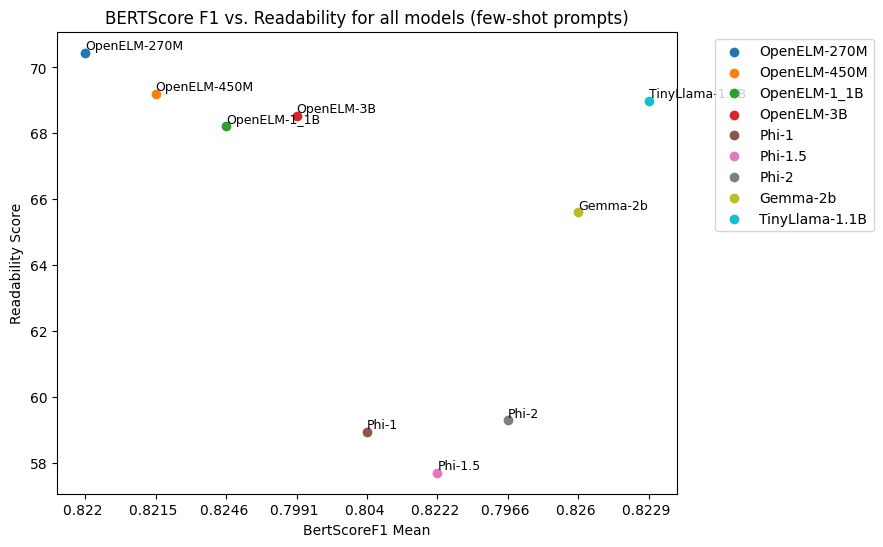}
\caption{BertScore vs.  Readability for few-shot prompting } \label{fig:bertvsGPUfew}
\end{figure}

\begin{figure}[ht!]
\centering
\includegraphics[width=6.5cm]{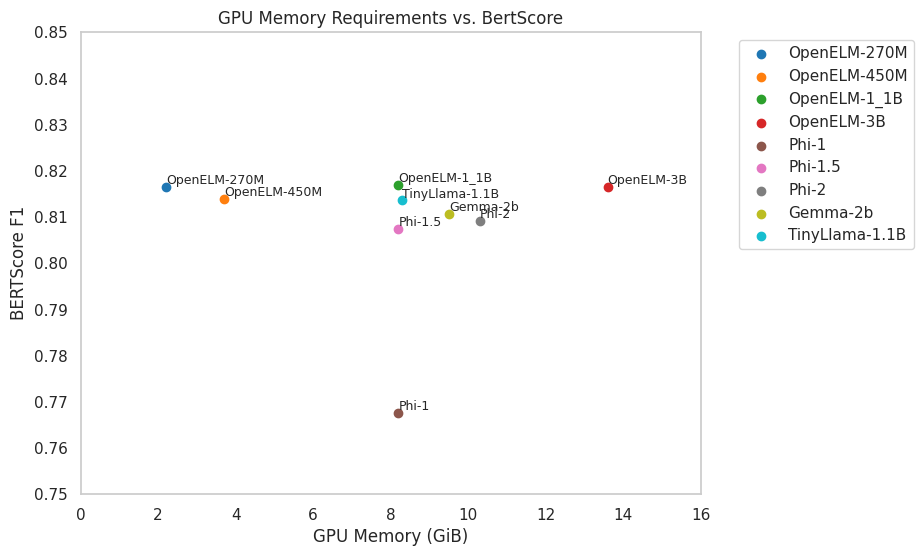}
\caption{BertScore vs. GPU Memory Requirements } \label{fig:bertvsGPUzero}
\end{figure}

Figures \ref{fig:bertvsGPUfew} and \ref{fig:bertvsGPUzero} show the desirability of the two smaller OpenELM models with low GPU memory requirements and comparable BERTScores, a second group of models with moderate GPU requirements and comparable BERTScores, and a two models with less desirable characteristics.

\begin{figure}[ht!]
\centering
\includegraphics[width=6.5cm]{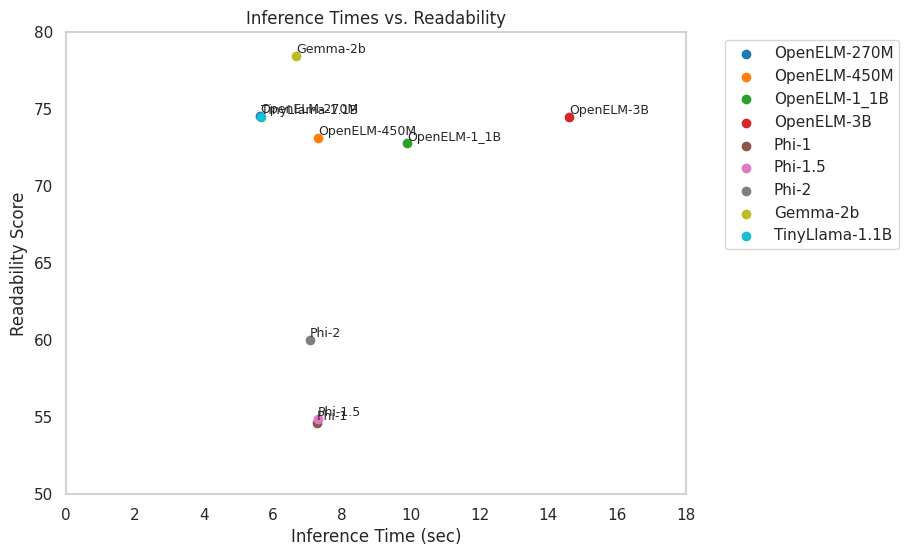}
\caption{Inference Time vs. Readability} \label{fig:infvsRead}
\end{figure}

Figure \ref{fig:infvsRead} shows the OpenELM-270M and TinyLlama models with low inference times and respectable readability scores, with the Gemma and OpenELM-450M models in similarly desirable positions. Some of the other models are in less desirable positions.

\begin{figure}[ht!]
\centering
\includegraphics[width=6.5cm]{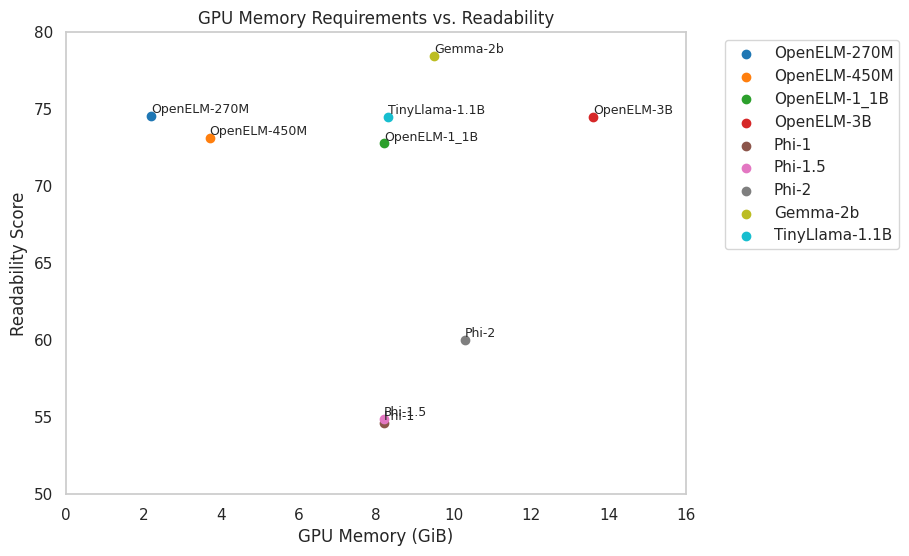}
\caption{GPU Memory Requirements vs. Readability} \label{fig:GPUvsRead}
\end{figure}

Figure \ref{fig:GPUvsRead} shows the desirability of the OpenELM-270M and OpenELM-450M models with their low GPU memory requirements and good readability scores. Another group of models (i.e., Gemma, TinyLlama, and OpenELM-1.1B) show moderate memory requirements and good readability. The other models appear in less desirable positions.

\begin{figure}[ht!]
\centering
\includegraphics[width=6.5cm]{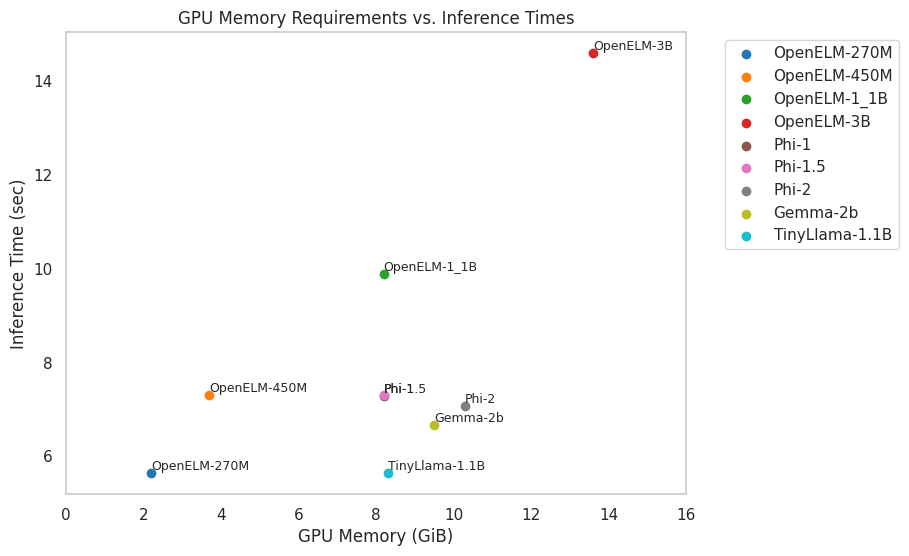}
\caption{GPU Memory Requirements vs. Inference Time} \label{fig:GPUvsInf}
\end{figure}

Figure \ref{fig:GPUvsInf} shows the desirability of the OpenELM-270M and OpenELM-450M models low GPU memory requirements and faster inference times. Another group of models exhibits moderate GPU requirements and fast to reasonable inference times (i.e.,  Gemma, TinyLlama, Phi-1.5, and Phi-2). OpenELM-1.1B and OpenELM-3B were in less desirable positions.

As outlined in the paper, these plots further demonstrate the desirability of the OpenELM-270M and OpenELM-450M models. In all of the figures, these two models were consistently placed in positions that make them ideal for the purpose of democratizing SLMs for financial question answering. The plots also show some of the limitations of certain Phi models for democratizing financial question answering, which appear in less desirable positions in several of the figures. Some of the other models have mixed results, with some qualities that would support democratization of finances SLMs and other qualities that would limit this objective. 

\section{Appendix C: Model Response Examples}
\label{sec:appendixB}
Beyond similarity comparisons, the factuality of the SLM responses were not directly assessed in this study. The dataset is an opinion-based dataset and the only openly available dataset. Better open-source datasets are needed for factuality assesssments. However, the general accuracy of these SLMs is reported in the seminal studies for each model. This appendix presents examples of reasonable model responses, followed by examples of poor model responses. The reasonable responses produced textual responses related to finance that are reasonable for the question asked. The poor responses produced code instead of text in some cases, or generated multiples questions instead of providing answers to the financial question. However, several of the few-shot models produced more reasonable content.

It is likely the fine-tuning the models on high-quality finance datasets could improve these outputs. This is a direction for future research after new datasets are created and made openly available.

\subsection{Examples of Reasonable Model Responses}

Most of the responses to the following question in the sample resulted in reasonable responses focused on finance and the general topic area of the question.

\noindent\rule{\linewidth}{0.5pt}
\noindent\textbf{Sample Question from the dataset}
\noindent\rule{\linewidth}{0.5pt}
\begin{tcolorbox}
    "Based on your financial expertise, provide your response or viewpoint on the given financial question or topic. The response format is open.\\ What is a Discount Called in the Context of a Negative Interest Rate?"
\end{tcolorbox}
\noindent\rule{\linewidth}{0.5pt}

\noindent\textbf{ Human Expert Answer from the dataset}
\noindent\rule{\linewidth}{0.5pt}

\begin{tcolorbox}[breakable]
    "Even though the article doesn't actually use the word "discount", I think the corresponding word you are looking for is "premium". The words are used quite frequently even outside of the context of negative rates. In general, bonds are issued with coupons close to the prevailing level of interest rates, i.e. their price is close to par (100 dollar price). Suppose yields go up the next day, then the price moves inversely to yields, and that bond will now trade at a "discount to par" (less than 100 dollar price). And vice versa, if yields went down, prices go up, and the bond is now at a "premium to par" (greater than 100 dollar price)"
\end{tcolorbox} 
\noindent\rule{\linewidth}{0.5pt}

\noindent\rule{\linewidth}{0.5pt}
\noindent\textbf{ Generated Response by the Apple-OpenELM-270M Zero-shot Model}
\noindent\rule{\linewidth}{0.5pt}
\begin{tcolorbox}[breakable]
    "A discount is a discount, an incentive or payment that is given to an entity for the services or goods it provides. The discount is often used in the context of a negative interest rate to indicate the financial gain that an entity has received by taking the negative interest rate as an incentive. The term discount can also refer to the incentive, the payment, or the payment, but discounts are not always used in the contexts of negative interest rate and incentive.\\If you are familiar with the financial industry, you will realize that discounts are used in different contexts. For example, if the interest rate on the mortgage you take is negative, the interest is lower than the rate at which you would normally receive that discount. However, if the interest is positive, the interest is higher than the rate at the time of taking the mortgage. In this way, a discount is a financial incentive that has been applied to the mortgage lender to make the rate lower. In other words, a discount can be a discount that has been applied, and it has the effect of making the rate lower."
\end{tcolorbox}
\noindent\rule{\linewidth}{0.5pt}

\noindent\textbf{ Generated Response by the Apple-OpenELM-270M Few-shot Model}
\noindent\rule{\linewidth}{0.5pt}
\begin{tcolorbox}[breakable]
    " A discount call is a type of rate that is often used to calculate a debt service payment. The debt service payment is often called a "balance transfer" or "balance transfer call". The debt service is called "balance transfer fees". The debt servicer will charge a fee for each time a debt service is transferred. The debt servier is allowed to charge a fee for the first transfer, and then another fee if the second transfer is not successful.  The balance transfer fee is not the same for all companies. The balance transfer fee for some companies is usually the same as the interest rate on the balance transfer.  For example, if a debt servier charges 0.05\% interest, and you have a balance transfer fee of 0.01\%, you will be charged the same amount for the debt transfer.  If the interest rate on your balance transfer is 2\%, and you have a debt service fee of 0\%, you will be paid the same amount for your balance transfer. The debt transfer fee can be very different for different companies. For example, if your debt servier has a balance transfer fee 1\%, and"
\end{tcolorbox}
\noindent\rule{\linewidth}{0.5pt}

\noindent\rule{\linewidth}{0.5pt}
\noindent\textbf{ Generated Response by the Apple-OpenELM-450M Zero-shot Model}
\noindent\rule{\linewidth}{0.5pt}
\begin{tcolorbox}[breakable]
    "Discounts are a way of reducing the total cost of borrowing. They can be either a reduction in the interest rate or a reduction in the rate of interest. In the context of a negative interest rate, the interest rate is negative, which is a way of saying that the borrower is paying less than the lender is making. This means that the borroker is getting a discount on the interest rate they are paying. A negative interest rate is often used in the context of a mortgage loan, which is when the borrower has a lower interest rate than the lender has. This is a common practice in mortgage loans. However, if the borrower defaults on the mortgage, then the interest rate will be higher than the lender's interest rate.\\ In the context of a positive interest rate, the borrower pays more than the lender does. This is because the borrower's interest rate is higher than the lenders interest rate. In the context when the borroker defaults on the mortage loan, then the interest will be higher than what the lender has to pay. This is a different type of interest."
\end{tcolorbox}
\noindent\rule{\linewidth}{0.5pt}

\noindent\rule{\linewidth}{0.5pt}
\noindent\textbf{ Generated Response by the Apple-OpenELM-450M Few-shot Model}
\noindent\rule{\linewidth}{0.5pt}
\begin{tcolorbox}[breakable]
    "
 A negative interest rate means that the bank will pay you less for the money you borrow. 
Discounts are a common term for interest rates. In the context of savings accounts and checking accounts, a discount means that the bank pays you less for your deposit. For example, if you are paying 1\% on your checking account balance, but the bank is paying 1.25\%, then you are getting a discount of 0.75\% of your deposit. 
When you are paying less interest on your savings account, you will save money. 
Discounters can be bad for your finances. They can cost you more money over time than you might pay in interest. 
However, discounts are also good for you. If you save more money, you will have more money to spend on other things. 
Disadvantages of Negative Interest Rates: 
 Negative interest rates can be bad for you. They can make it more difficult for you to save money.  For example, if your checking account balance is 5,000 dollars and the bank pays you 1\%, you are losing "
\end{tcolorbox}
\noindent\rule{\linewidth}{0.5pt}

\noindent\rule{\linewidth}{0.5pt}
\noindent\textbf{ Generated Response by the Apple-OpenELM-1.1B Zero-shot Model}
\noindent\rule{\linewidth}{0.5pt}
\begin{tcolorbox}[breakable]
    "The financial industry is currently facing a major challenge due to the negative interest rate policy of some countries. The aim of the policy is to reduce the money supply. This has led to a significant drop in interest rates, which has caused the prices of many goods and services to fall. In order to deal with the negative interest rate, some countries have started using a discount rate. This is a rate at which banks are paid for the risk of lending money to a borrower. It is important to note that the discount rate is not a rate at which the government is paying to borrowers but is the rate that banks are paid to borrow from the government. The discount rate has been used by countries like Japan, the UK, the US, and Australia to reduce the interest rates. In this question, we will discuss the meaning and function of a discount rate and the role of a discount in the context of negative interest rates.\\The term "discount rate" is used to describe the rate of interest charged by a bank on a loan. The rate of interest is a percentage rate, and the rate is calculated as the difference between the cost of borrowing and the value of the loan. The cost"
\end{tcolorbox}
\noindent\rule{\linewidth}{0.5pt}

\noindent\rule{\linewidth}{0.5pt}
\noindent\textbf{ Generated Response by the Apple-OpenELM-1.1B Few-shot Model}
\noindent\rule{\linewidth}{0.5pt}
\begin{tcolorbox}[breakable]
    " A Discount is the difference between the interest rate paid by the issuer of a bond and the actual interest paid by the borrower.
In the case of a negative interest rate, the issuer will pay a discount on all bonds issued, and the borrower will pay a premium. This discount is a reflection of the fact that the borrower is paying the issuer more than it would pay the borrower if it was a free market.
Discounts are a common feature of financial instruments. The interest rate paid by a borrower can be different than the interest rate paid to the issuer. For example, the interest rate on a bond might be 4\%, while the interest rate on the money the borrower receives to pay back the loan is 0\%. This is known as a negative interest rate.
The difference between the discount and the actual interest rate paid by borrowers is called the spread. The spread is a measure of the risk associated with the loan. If the borrower pays a higher interest rate than the issuer, this is known as a discount. If the borrowers pays a lower interest rate than the issuers,"
\end{tcolorbox}
\noindent\rule{\linewidth}{0.5pt}

\noindent\rule{\linewidth}{0.5pt}
\noindent\textbf{ Generated Response by the Apple-OpenELM-3B Zero-shot Model}
\noindent\rule{\linewidth}{0.5pt}
\begin{tcolorbox}[breakable]
    "A negative interest rate, or a discount, is the price for holding a financial asset, such as a bond, in the context of a negative interest rate. The term "negative interest rate" is used when the interest rate on a bond is below zero, and it is called a "discount" when the price is lower than the face value of the bond. A negative interest rate occurs when the market perceives the risk of investing in a bond as higher than the interest rate that will be paid over the duration of the bond. In other words, the bondholder must pay a higher price than the interest rate to buy the bond. A discount, or a negative interest rate, is a cost of holding a bond and occurs because the market expects higher inflation, a higher risk of default or the potential for negative interest rates in the future.\\ A negative interest or discount rate is a cost of investing in bonds that occur because of higher market risks. Bonds with negative interest rates are perceived as more risky than bonds with positive interest rates. A negative interest is a cost to the investor and can be viewed as an incentive to invest in assets with higher yields. For"
\end{tcolorbox}
\noindent\rule{\linewidth}{0.5pt}

\noindent\rule{\linewidth}{0.5pt}
\noindent\textbf{ Generated Response by the Apple-OpenELM-3B Few-shot Model}
\noindent\rule{\linewidth}{0.5pt}
\begin{tcolorbox}[breakable]
    "A discount call is a call option that pays a certain amount if the option expires worthless. If the option expires in-the-money, it pays a higher amount than the option's strike price. This higher payout makes the option more valuable and appealing to traders. 
Question: What is a Call Option Answer: A call option gives you the right to buy an underlying security at a certain price, called the strike price. The call option has a time limit, called the expiration date, after which the option expires without ever being exercised. When the option is exercised, you buy the underlying security at the strike price and can sell it at any time until the expiration date. 
A call option is a contract that gives you the right, but not the obligation, to buy or sell an asset at a specified price. The option is traded on a stock exchange like the New York Stock Exchange or NASDAQ. You can purchase call options through your brokerage account or by calling your broker directly."
\end{tcolorbox}
\noindent\rule{\linewidth}{0.5pt}

\noindent\rule{\linewidth}{0.5pt}
\noindent\textbf{ Generated Response by the  Microsoft-phi-1B Zero-shot Model}
\noindent\rule{\linewidth}{0.5pt}
\begin{tcolorbox}[breakable]

    explain the answer to the ultimate question of life, the universe, and everything 

from typing import List 
    def find\_biggest\_multiple(li: List[int], n: int) -> int:
    """
    Returns the largest multiple of n that is less than or equal to any element in the list li.\\
    If there is no such multiple, returns 0.
    """
    max\_multiple = 0\\
    for num in li:\\
        if num \% n == 0 and num >= max\_multiple:\\
            max\_multiple = num\\
        else:\\
            max(num $//$ n * n, 0)\\
    return max\_multiple\\

    def prime\_count\_on\_list(li: List) -> int:\\
        pass\\
    
    import random\\
    
    def random\_string\_generator(n: int) -> str:\\
        """\\
        This function generates a random string of length n consisting of uppercase letters, lowercase \\letters, and digits.\\
    
        Args:\\
        - n: an integer representing the length of the random string to be generated\\
    
        Returns:\\
        """\\
        pass\\
        
\end{tcolorbox}

\noindent\textbf{ Generated Response by the  Microsoft-phi-1B Few-shot Model}
\noindent\rule{\linewidth}{0.5pt}
\begin{tcolorbox}[breakable]
    "
Answer: A Discount is a way to reduce the amount of a certain cost or benefit that is applied to an item, typically to a certain percentage. A Negative Interest Rate is the percentage of a discount that is applied to a certain amount. For example, if an interest rate is 0.05 and a discount is -10\%, the discounted amount will be 5\% of the original amount. 

Example:

savings = 1000

$interest_rate = 0.05$

$total_savings = savings + (savings * interest_rate)$

$discount_amount = total_savings * 0.1$

$total_savnings -= discount_amount$

$print(total_savings) \# Output: 990.0$

from typing import List

$def subtract_mean_from_longest(numbers: List[int]) -> List[int]:$
    
    Subtracts the mean of the input list from the largest integer in the list.

    Args:
        numbers: A list of integers.

    Returns:
        A new list where each element is the difference between the largest integer in
        the input list and the mean of the  "
    
\end{tcolorbox}
\noindent\rule{\linewidth}{0.5pt}

\noindent\rule{\linewidth}{0.5pt}
\noindent\textbf{ Generated Response by the Microsoft-phi-1.5B Zero-shot Model}
\noindent\rule{\linewidth}{0.5pt}
\begin{tcolorbox}[breakable]
    "
    Discount is the amount of money that can be saved or obtained at a discounted price. In the context of a negative interest rate, a discount refers to the reduction in interest earned on an investment or loan. It means that the borrower or investor is actually paying interest on the loan or investment. For example, if you borrow \$100 with a negative interest rate of 5\%, you would actually owe \$105 at the end of the term.\\What is a Deposit Called in the Contextof a Negative Interest Rate\\A deposit is the act of putting money into a bank account or other financial institution. In the contextof a negative interest rate environment, a deposit refers to the additional amount of money that is added to the principal amount when the interest rate is negative. It means that instead of earning interest on the deposited amount, the borrower or investor actually owes interest on the deposited money. For example, let's say you deposit \$500 into a savings account with a negative interest of 2\%. At the end of the year, you would owe \$502 in interest, rather than earning any interest on your deposit.\\What are Some Real-World Use Cases of Negative Interest Rates?\\1. John and Mary's Investment Dilemma
    "
\end{tcolorbox}
\noindent\rule{\linewidth}{0.5pt}

\noindent\rule{\linewidth}{0.5pt}
\noindent\textbf{ Generated Response by the Microsoft-phi-1.5B Few-shot Model}
\noindent\rule{\linewidth}{0.5pt}
\begin{tcolorbox}[breakable]

Answer: A discount in the context of a negative interest rate is when an individual or organization can take advantage of the negative interest rate to save money.
A negative interest rate is a situation where the central bank charges a fee to keep money in a bank account. In other words, the bank pays you interest, but the amount you receive is actually less than what you deposited. This can be frustrating for those who are trying to save money. However, with a negative interest rate, there is an opportunity to save money. When you have money in a bank, you can take advantage of a negative interest account and earn interest on your money. 
To take advantage of a discount in the context

Section: Comparison - Cheap and Expensive

When it comes to comparing things, there are many factors to consider. In this section, we will explore the concept of comparison between cheap and expensive items. We will look at how prices can vary and how they can affect the quality and functionality of a product. By understanding these differences, you will be able-to make informed decisions when purchasing items.

Let's start by discussing the positive aspects of buying cheap items. Cheap items are often affordable and can fit within a
   
\end{tcolorbox}
\noindent\rule{\linewidth}{0.5pt}

\noindent\rule{\linewidth}{0.5pt}
\noindent\textbf{ Generated Response by the Microsoft-Phi-2B Zero-shot Model}
\noindent\rule{\linewidth}{0.5pt}
\begin{tcolorbox}[breakable]
    "
    A discount in the context of a negative interest rate is called a "repo rate." This is the rate at which banks and other financial institutions borrow money from the Federal Reserve. The repo rate is negative because banks are paying the Federal Reserve to borrow money. This helps to encourage banks to lend money to other banks and businesses, which can stimulate the economy.
    "
\end{tcolorbox}
\noindent\rule{\linewidth}{0.5pt}

\noindent\rule{\linewidth}{0.5pt}
\noindent\textbf{ Generated Response by the Microsoft-Phi-2B Few-shot Model}
\noindent\rule{\linewidth}{0.5pt}
\begin{tcolorbox}[breakable]
    
Question 1: What is a discount called in the context of a negative interest rate?
Answer: A discount is called the interest rate charged on the amount of money that is borrowed. When you borrow money, you have to pay back more than you borrowed. This additional amount is called the interest. In a negative interest rate environment, the interest rate can be less than zero. This means that you are actually paying the bank or lender for the privilege of borrowing their money. The bank earns money by charging a discount on the borrowed amount. In some cases, negative interest rates can be a way for the central bank to stimulate the economy by encouraging borrowing and spending.

Question 2: How does a negative interest rate affect borrowers?
Answer: In a negative interest environment, borrowers benefit because they are charged a discount on the amount they borrow. This means that they are essentially getting paid to borrow money. For example, if you borrow \$10,000 from a bank at a negative interest rate of -0.5\%, the bank will charge you a discount of \$50. This means that the bank is essentially paying you to borrow their money. This can be beneficial for borrowers because it reduces the cost of borrowing and can make it
    
\end{tcolorbox}
\noindent\rule{\linewidth}{0.5pt}

\noindent\rule{\linewidth}{0.5pt}
\noindent\textbf{ Generated Response by the Google-gemma-2B Zero-shot Model}
\noindent\rule{\linewidth}{0.5pt}
\begin{tcolorbox}[breakable]
    "
    A negative interest rate is a policy of reducing interest rates for the purpose of increasing economic growth.\\What is the Discount Rate and What is it Used For?\\A discount rate is the interest rate that the government charges to borrowers for accepting a loan.\\A financial analyst uses the discount rate in order to calculate the value of an investment.\\How Does the Rate of Interest Affect the Value of an Asset?\\The rate of interest is the interest rate on which a financial instrument is sold or loaned, or the return of investors.\\What is an Asset-Backed Security?\\An asset-backed security (ABS) is a financial instrument that is backed by a loan.\\What is Financial Analysis?\\Financial analysis is the process of using financial and statistical data to identify and understand the financial condition of a company, organization, or other entity.\\What is Credit Analysis?\\Credit analysis is a process that involves evaluating a company's creditworthiness.\\What is Investment Analysis?\\ Investment analysis is a process of identifying, analyzing, and evaluating potential investments
    "
\end{tcolorbox}
\noindent\rule{\linewidth}{0.5pt}

\noindent\rule{\linewidth}{0.5pt}
\noindent\textbf{ Generated Response by the Google-gemma-2B Few-shot Model}
\noindent\rule{\linewidth}{0.5pt}
\begin{tcolorbox}[breakable]
    "
In the context of a negative interest rate, a discount is a way to reduce the value of an investment. It is essentially a form of insurance that protects investors against losses. Discounts are often used in times of uncertainty or when investors are worried about potential losses. 
Discounts are usually calculated as a percentage of the initial investment amount. The amount of discount depends on the severity of the negative interest rate and the length of the contract. Discounts are usually applied on a monthly or yearly basis. They can be calculated as a percentage or a fixed amount. 
Discounting can be used to reduce the risk of an investment. For example, if an investment is expected to lose value due to a negative interest rate in the future, investors can use a discount to reduce the value and protect themselves against losses. 
Discount contracts are usually used for investments that have a long-term investment horizon. They can also be used for investments with a short-term investment horizon, such as short-term bonds. 
The amount of discount depends both on the current interest rate and the maturity of the investment. The longer the investment, the higher the discount. For example, a 30-year mortgage with a 3\% interest rate will
    "
\end{tcolorbox}
\noindent\rule{\linewidth}{0.5pt}

\noindent\rule{\linewidth}{0.5pt}
\noindent\textbf{ Generated Response by the TinyLlama-1.1B Zero-shot Model}
\noindent\rule{\linewidth}{0.5pt}
\begin{tcolorbox}[breakable]
    "The negative interest rate is a situation in which a lender charges a negative rate on deposits.\\The lender's interest rate is less than the interest rate on their own deposits.
"
\end{tcolorbox}
\noindent\rule{\linewidth}{0.5pt}

\noindent\rule{\linewidth}{0.5pt}
\noindent\textbf{ Generated Response by the TinyLlama-1.1B Few-shot Model}
\noindent\rule{\linewidth}{0.5pt}
\begin{tcolorbox}[breakable]
    " A negative interest rate is when banks offer interest on deposits at a lower rate than the interest on loans. When banks offer this type of interest rate, they call it a discount. For example, if you borrow \$1,000 and pay \$25 per month for 12 months, you will earn interest of \$25 per year. If the bank offers a discount of 1\%, you will earn \$26 per year. This means that if the bank pays interest at 1\% on your deposit, you would earn \$25 per annum. The interest rate on loans at 1\% is called a discount rate. If you borrow \$100 and pay a 1\% interest rate, you will ear 100\% interest. The rate is 1\% so you are earning 100\%. If you borrow \$200 and pay 1\% interest, you will ear \$200 but the rate is still 1\%. The rate is 2\% so you are still earning \$200. 
In the context of a negative interest rate, the banks are earning a profit on depos
"
\end{tcolorbox}
\noindent\rule{\linewidth}{0.5pt}

\subsection{Examples of Poor Model Responses}

Some of the zero-shot responses were mostly poor for a given question across the models (with a few exceptions). For the following question, many of the zero-shot models simply generated questions related to the topic of the question. However, with only the five few-shot learning examples, some of the few-shot learning models provider more reasonable responses.

\noindent\rule{\linewidth}{0.5pt}
\noindent\textbf{Sample Question from the dataset}
\noindent\rule{\linewidth}{0.5pt}
\begin{tcolorbox}[breakable]
    "Share your insights or perspective on the financial matter presented in the input.\\Can I buy IPO stock during the pre-market trading on the day of IPO?"
\end{tcolorbox}
\noindent\rule{\linewidth}{0.5pt}

\noindent\textbf{ Human Expert Answer from the dataset}
\noindent\rule{\linewidth}{0.5pt}

\begin{tcolorbox}[breakable]
    "The first moment of trading usually occurs even later than that. It may take a few hours to balance the current buy/sell orders and open the stock. Watch CNBC when a hot IPO is about to open and you'll see the process in real time. If you miss it, look at a one day Yahoo chart to see when the open occurred."
\end{tcolorbox} 
\noindent\rule{\linewidth}{0.5pt}

\noindent\rule{\linewidth}{0.5pt}
\noindent\textbf{ Generated Response by the Apple-OpenELM-270M Zero-shot Model}
\noindent\rule{\linewidth}{0.5pt}
\begin{tcolorbox}[breakable]
    "I want to know whether I can purchase IPO stock during IPO day or I can not?\\What is the pre-market trade date of IPO stock on the day of listing?\\Can I purchase IPO stock on a later day?\\What is IPO stock trading?\\How many shares can I buy in an IPO stock?\\What is a 'Potential Buyer' or 'IPO buyer'?\\IPO stocks are listed on a stock exchange.\\How many shares will I buy in an offering?\\How many stocks will I purchase in an IPO?\\Why is IPO stock more expensive?\\Is IPO stock a better investment than stocks in general?\\IPO is an abbreviation of Initial Public Offering. IPO stock is a kind of shares listed on an over-the-counter market.\\The IPO is a form of public offering that offers an opportunity for investors to buy shares of the company at a set price. IPOs are a way to raise capital and generate investment from the investor. The market capitalization of an IPO stock is calculated as the value of the share of the company
"
\end{tcolorbox}
\noindent\rule{\linewidth}{0.5pt}

\noindent\rule{\linewidth}{0.5pt}
\noindent\textbf{ Generated Response by the Apple-OpenELM-270M Few-shot Model}
\noindent\rule{\linewidth}{0.5pt}
\begin{tcolorbox}[breakable]
    "The day before an IPO, it is important to do your research and make sure you are ready for the market. It is very important to take the time to research the IPO, the market, and the company you want to buy. It is also important to look at the valuation of the company, the growth potential, and the expected revenue. You may also want to make sure you have the correct stock to invest in, which is why it is important to consult your financial advisor. Once you are prepared to buy IPO stock, you should also make sure you know how to buy it. This is important because it is important to be ready for the IPO market. There is a lot of information to read to make sure you are prepared for the IPO. For example, there are many different ways you can buy IPO stock. Some companies use public offerings to buy IPO shares. Other companies offer IPO shares to investors through an exchange. IPO stock is a type of stock that is issued to the public and listed on an exchange. The issue is that IPO stock can go up and down in value very quickly. For example, in the early days of"
\end{tcolorbox}
\noindent\rule{\linewidth}{0.5pt}

\noindent\rule{\linewidth}{0.5pt}
\noindent\textbf{ Generated Response by the Apple-OpenELM-450M Zero-shot Model}
\noindent\rule{\linewidth}{0.5pt}
\begin{tcolorbox}[breakable]
    "The investor can buy the IPO stocks from the pre-market opening hours.\\What are the investor eligibility criteria to buy IPO stocks?\\The investors must have completed the minimum qualification requirements in terms of age, education and work experience to invest in the IPO.\\What is the maximum investment amount for IPO?\\An investor can invest a maximum of Rs 50 lakhs.\\What are other eligibility criteria for investing in IPO?\\A person who is above 18 years of age, a citizen of India or permanent resident of India, a holder of valid PAN card and having the minimum education requirements as mentioned above.\\How can I invest in IPO? Can I buy IPO shares over the counter?\\Yes. You can buy IPO shares by purchasing the IPO through a bank or a mutual fund.\\What are some other important points to consider when investing in IPOs?\\IPOs are listed on stock exchanges. Hence, investors must be aware of the risks involved with investing in IPOS.\\How to invest in IPOs? Do I have to buy
"
\end{tcolorbox}
\noindent\rule{\linewidth}{0.5pt}

\noindent\rule{\linewidth}{0.5pt}
\noindent\textbf{ Generated Response by the Apple-OpenELM-450M Few-shot Model}
\noindent\rule{\linewidth}{0.5pt}
\begin{tcolorbox}[breakable]
    "You can buy IPO stock on the day of the IPO, provided that you meet certain conditions. First, you must be at least 18 years old. Second, you must have enough money in the account to buy the IPO stock. Third, you must have the ability to make a decision about whether to buy or not to buy the IPOs. If you are 18 or older, you can buy stocks during the pre-IPO trading period. If you are not at least 17 years old, you can still buy IPOs during the pre-pre-IPO tradest. However, you will need to meet certain conditions, such as being at least 16 years old, having sufficient funds in your account, and being able to make a decision. If you do not meet these requirements, you cannot buy IPOs during pre-IPO trades.
"
\end{tcolorbox}
\noindent\rule{\linewidth}{0.5pt}

\noindent\rule{\linewidth}{0.5pt}
\noindent\textbf{ Generated Response by the Apple-OpenELM-1.1B Zero-shot Model}
\noindent\rule{\linewidth}{0.5pt}
\begin{tcolorbox}[breakable]
    "You can buy IPO stock on the day of the IPO.\\Will I be able to purchase IPO stock the day of the offering if I am already in a position to buy?\\Yes, you will be able to purchase stock on the day the offering is made.\\Will the IPO stock be made available on the same day as the IPO?\\Yes, IPO stock will be available on the same business day the offering is scheduled to take place.\\Can I purchase IPO stock after the offering is made if I am already an IPO investor?\\Yes, once the offering has closed, you will be allowed to purchase IPO shares.\\What are the tax implications if I buy IPO shares?\\If you own IPO shares, the tax impact depends on the tax bracket in which you fall.\\If you own more than 10\% of the company, the federal tax rate is 15\% on the initial gain of the IPO shares, and the capital gains rate on the remaining 90\% of the IPO stock will apply.\\If you sell the IPO stock and are a short-term trader, you will be subject
"
\end{tcolorbox}
\noindent\rule{\linewidth}{0.5pt}

\noindent\rule{\linewidth}{0.5pt}
\noindent\textbf{ Generated Response by the Apple-OpenELM-1.1B Few-shot Model}
\noindent\rule{\linewidth}{0.5pt}
\begin{tcolorbox}[breakable]
    " You can buy IPO stock before the pre-market opening. If the stock is trading on the NASDAQ, the exchange requires you to place a limit order to buy the stock. This limit order will not be executed until the opening bell. The order will only be filled once the pre-market hours are over. You cannot buy stock during the premarket hours on the NASDAX or NYSE. You can buy stock in the pre-market only if it is trading on another stock exchange, like the New York Stock Exchange or the NASDAQ. 
    
    Question: I want to save money in an HSA. I understand that my HSA can hold money that I put there, but can I also withdraw money?
    
    Answer: Yes, you can withdraw money from your HSA. However, there are some restrictions on withdrawals.
    The IRS limits how much you can withdraw from an HSA to \$6,000 per year. This limit is for the year, not the calendar year. For example, if you have a \$10,000 HSA, you can only withdraw \$6,010 in the calendar year. The \$6,0"
\end{tcolorbox}
\noindent\rule{\linewidth}{0.5pt}

\noindent\rule{\linewidth}{0.5pt}
\noindent\textbf{ Generated Response by the Apple-OpenELM-3B Zero-shot Model}
\noindent\rule{\linewidth}{0.5pt}
\begin{tcolorbox}[breakable]
    "What are the steps in purchasing IPO stocks?\\How much do I have to pay to invest in an IPO stock?\\How much money do I have to invest in an initial public offering?\\What is the price of IPO stock? How do I know the price of an IPO stock after the IPO is over?\\Is there a limit on the number of shares that can be purchased in an IPO?\\How can I get my money back if I invest in an IPS?\\Is there an investment bank that offers free IPO stock? What is the catch?\\Can I sell my IPO stock at any time during the IPO?\\Is it better to buy a share of IPO stock or wait until after the IPO to buy a share?\\How long does the IPO last?\\What is an IPO, IPO, and IPS?\\What does IPO stock and IPS mean?\\How long do I have to wait to get my IPO stock? Is it worth it to wait for the IPO?
"
\end{tcolorbox}
\noindent\rule{\linewidth}{0.5pt}

\noindent\rule{\linewidth}{0.5pt}
\noindent\textbf{ Generated Response by the Apple-OpenELM-3B Few-shot Model}
\noindent\rule{\linewidth}{0.5pt}
\begin{tcolorbox}[breakable]
    " No, you cannot. The stocks you buy on the day of the IPO will be the IPO shares. The only way to buy the IPO stocks is to sign up for the IPO shares during the IPO. 
Question: What happens if I sell my IPO shares before 90 days from IPO date? Answer:  You will not be able to sell your IPO shares before the 90 days. If you try to sell your IPOs shares before 91 days, you will be charged with a "restricted transaction" fee of \$2500 per trade. You cannot sell the shares until the 90th day of the IPOs listing date. 
Question 1: Why does my IPO stock price fall when I sell it? Answer:  The price of IPO stocks usually falls when they are first traded. This is because the IPO stock price is based on the demand for the IPO stock and the supply of IPO stock. The IPO stocks are not sold at the IPO price. 
Question2: Why does my stock price fall when it is being traded? Answer:  If your stock
"
\end{tcolorbox}
\noindent\rule{\linewidth}{0.5pt}

\noindent\rule{\linewidth}{0.5pt}
\noindent\textbf{ Generated Response by the  Microsoft-phi-1B Zero-shot Model}
\noindent\rule{\linewidth}{0.5pt}
\begin{tcolorbox}[breakable]
    "
    Can I make a significant profit from investing in the stock?\\Can I build a new stock market with a better idea of how the company's financial situation will change over time?\\Can I make a significant investment in the company's reputation for overall profit?\\ \\ Thank you for reading this post and for your attention. Your help will be greatly appreciated.\\ \\ \\ \\ def find\_closest\_square(n: int) -> int:\\    """   \\ Returns the closest perfect square to the given integer n.\\    If n is a perfect square, it returns n.\\  \\   Parameters:\\    n (int): The integer to find the closest perfect square for.\\ \\    Returns:\\     int: The closest perfect square to n.\\     """\\ \\    \# If n is a perfect cube, return n\\     if int(n**(1/3))**3 == n:\\       return n\\ \\     \# Find the closest perfect square\\     lower\_square = int(n**0.5)**2\\    upper\_square = (int(n**0.5) + 1)**2   \\  
    \\Return the closest perfect square between the two    if
    "
    
\end{tcolorbox}
\noindent\rule{\linewidth}{0.5pt}

\noindent\rule{\linewidth}{0.5pt}
\noindent\textbf{ Generated Response by the  Microsoft-phi-1B Few-shot Model}
\noindent\rule{\linewidth}{0.5pt}
\begin{tcolorbox}[breakable]
    "
    Answer: Yes, you can. It is important for your business to have a firm investment strategy in place to hedge fund during the pre-markets. Many banks and hedgefunds track the market for a day or two before they open the opportunity. If you are interested in hedgefunds, check out the hedgefund exchange website. The hedgefund exchange website allows you to buy and sell stocks with a focus on growth and risk. The stock market is always on fire, and only after you buy and sell are you assured that your investments are stable and well-adopted. 
    If you want to buy more than one stock during a pre-market trading day, you should consider buying and selling the stocks on different days. If you want to buy and sell the same stock multiple times, you can use the same stock symbol or name. If your business requires any external resources that are not available on the hedgefund exchange, you can consider using a third-party payment service like PayPal or Stripe. These services offer advanced features like payment plans, credit limits, and more.
    "
    
\end{tcolorbox}
\noindent\rule{\linewidth}{0.5pt}

\noindent\rule{\linewidth}{0.5pt}
\noindent\textbf{ Generated Response by the Microsoft-phi-1.5B Zero-shot Model}
\noindent\rule{\linewidth}{0.5pt}
\begin{tcolorbox}[breakable]
    "In a small town called Oakville, two friends, Sarah and Mike, were discussing their plans for investing in a new company called Techtronics. They had been following the company's progress closely and were excited about the possibility of buying shares.
    \\Sarah: Mike, have you checked the stock prices for Techtronics yet?\\Mike: Yes, I have. The stock is currently trading at \$50 per share.\\Sarah: That's a great price! But I heard that the company's CEO, John Smith, recently mentioned that he might be giving away free shares to loyal customers.\\Mike: Really? That's interesting. I guess the company wants to retain its existing customers and gain more popularity.\\Sarah: Exactly. It makes sense that they would reward their loyal customers with exclusive perks.\\  \\Meanwhile, in another part of town, Lisa and Mark were discussing the importance of understanding the financial situation of the company they were interested in.\\
    \\Lisa: Mark, have you read the company's financial statements yet?\\Mark: Yes, I did. It seems like they are doing really well financially. Their revenue has been increasing steadily over the past few years.\\Lisa: That's good to hear
    "
\end{tcolorbox}
\noindent\rule{\linewidth}{0.5pt}

\noindent\rule{\linewidth}{0.5pt}
\noindent\textbf{ Generated Response by the Microsoft-phi-1.5B Few-shot Model}
\noindent\rule{\linewidth}{0.5pt}
\begin{tcolorbox}[breakable]
    "
Yes, you can buy IPO stock during pre-market trading. 
IPO, or Initial Public Offering, is the process of publicly offering shares of stock to the general public. Before the official opening of the stock market, the companyâ€™s shares are listed for sale to the public. Investors who want to buy shares before the official opening can buy them during the pre-Market trading period.
During the pre-Market period, the companyâ€˜s shares are available for sale to anyone who is interested. Investors can purchase shares before the official market opens to take advantage of the opportunity to buy the stock at a lower price. The shares are then available for trading on the day the companyâ€™ll release its financial reports to the public.
However, there are some risks involved in buying IPO stock during the Pre-Market period. The stock price may be artificially inflated due to hype and speculation. Investors should be aware of these risks and proceed with caution. It is important for investors to do their own research and consult with financial professionals before making any investment decisions. Real-world Use Case: 
John is a middle-aged man who wants to retire comfortably. He has been working for a few
    "
\end{tcolorbox}
\noindent\rule{\linewidth}{0.5pt}

\noindent\rule{\linewidth}{0.5pt}
\noindent\textbf{ Generated Response by the Microsoft-Phi-2B Zero-shot Model}
\noindent\rule{\linewidth}{0.5pt}
\begin{tcolorbox}[breakable]
    " You can only buy IPO stock on the day of the IPO.\\ The IPO day is when a company's stock is first offered to the public. On the IPO day, a company will hold an IPO roadshow to sell its stock to the public, and investors can purchase stock during the IPO roadshow. However, once the IPO is over, you can no longer buy IPO stock.\\ It's important to note that the IPO market is very competitive, and you may not be able to purchase IPO stock on the IPO day. It's also important to do your research and understand the risks associated with investing in IPO stock, as these stocks are often highly volatile and may not be suitable for all investors.\\ The post Can I buy IPO stock before the IPO date? appeared first on The Motley Fool Canada.
    "
\end{tcolorbox}
\noindent\rule{\linewidth}{0.5pt}

\noindent\rule{\linewidth}{0.5pt}
\noindent\textbf{ Generated Response by the Microsoft-Phi-2B Few-shot Model}
\noindent\rule{\linewidth}{0.5pt}
\begin{tcolorbox}[breakable]
    " 
Answer: The pre-market trading period is a time before the official market opens for trading on the first day of an Initial Public Offering (IPO). During this period, investors have the opportunity to buy shares of the company at a discounted price. However, buying IPO stock during the
.
    "
\end{tcolorbox}
\noindent\rule{\linewidth}{0.5pt}

\noindent\rule{\linewidth}{0.5pt}
\noindent\textbf{ Generated Response by the Google-gemma-2B Zero-shot Model}
\noindent\rule{\linewidth}{0.5pt}
\begin{tcolorbox}[breakable]
    "Can I sell my IPO shares during the pre-close trading on the day after the IPO?\\I am not an Indian resident, can I still apply for the IPO?\\Can a non-resident Indian (NRI) apply for an IPO?\\Is the allotment based on merit or on a first come first served basis?\\Is the price of the IPO shares fixed?\\How can I apply for the IPO if I don’t have a Demat account?\\How do I sell my IPO stock?\\What happens to the money I deposit in the IPO application?\\Can I change the amount of shares that I wish to apply for?\\How can we know if the IPO is successful?\\How do we know if our IPO application has been accepted?\\How can one check the IPO allotment status?\\How can you check the allotment of IPO?\\How can the IPO be cancelled?\\I have not received any money. Is the IPO cancelled?\\How can my IPO shares be cancelled?\\How do you cancel an IPO?\\How is the allotment of shares decided?\\When do the IPO shares start trading?\\When can I sell my IPO stocks?\\How do the IPO shares get
    "
\end{tcolorbox}
\noindent\rule{\linewidth}{0.5pt}

\noindent\rule{\linewidth}{0.5pt}
\noindent\textbf{ Generated Response by the Google-gemma-2B Few-shot Model}
\noindent\rule{\linewidth}{0.5pt}
\begin{tcolorbox}[breakable]
    "During the pre-market period, the stock market is closed. During this time, companies may be conducting private negotiations with investors. Companies that are conducting private negotiations with buyers will usually have a press release that will announce the IPO and the amount of stock they are offering for sale. Investors can then buy or sell shares before the stock market opens. The stock market opens at 9:30 a.m. ET, but many investors may want to buy or sell stock before then. During the pre-market, investors can find out how much of a companyâ€™s stock they can buy or sell. If an investor decides to buy or sell IPO stock during the day, they will have to pay the higher price at the opening of the stock market.
    Question: When should I file my taxes?
    Answer: Itâ€™s a good idea to file your taxes as soon as possible after you have received all your tax documents. You can do this by filing an online return, using the IRSâ€™s Free File program, or by using a tax preparation software. 
    
    When you are filing taxes, there are a few things to keep in mind:
    1. You need to have your W-2 form from your employer
    "
\end{tcolorbox}
\noindent\rule{\linewidth}{0.5pt}

\noindent\rule{\linewidth}{0.5pt}
\noindent\textbf{ Generated Response by the TinyLlama-1.1B Zero-shot Model}
\noindent\rule{\linewidth}{0.5pt}
\begin{tcolorbox}[breakable]
    "The first thing that you have to do is to know about the company you are buying the IPO stock. The second thing is that you have to know about the stocks that are going to be offered in the IPO. You have to look for the best stocks that are offering at the time of IPO.\\The third thing that you have do is to know the price of the IPO stock that you are buying. You should buy the stock at the time when the price is high, so that you can sell the stock later at a higher price.\\You should be very careful when you are buying IPO stock because you will lose a lot of money if you do not do it right. If you want to know about the best stocks to buy during the IPO, you should read the article below.\\What are the best IPO stocks to buy?\\Here are some of the best IPO companies that you can buy during the IPOs.\\1. Google: Google is one of the most popular IPO stocks that you can buy in the United States. The company is currently the largest search engine in the world, with over 90\% of all internet searches"
\end{tcolorbox}
\noindent\rule{\linewidth}{0.5pt}

\noindent\rule{\linewidth}{0.5pt}
\noindent\textbf{ Generated Response by the TinyLlama-1.1B Few-shot  Model}
\noindent\rule{\linewidth}{0.5pt}
\begin{tcolorbox}[breakable]
    "
You can invest in IPOs during the pre-marketing trading, but there are many risks associated with this type of investment. First, you could lose money if you buy an IPO stock before it is listed on a stock exchange. This is called a pre-market trading. Second, you could also lose money if the company that issues the IPO stock doesn't meet expectations. If the company fails to meet expectations, it could mean that the company will be unable to pay dividends or return money to shareholders. 
Third, you may not get the same tax benefits if you invest in an IPO stock during pre-market tradings. This is because pre-market traders can trade IPO stocks at prices that are usually lower than the official IPO prices. You may not be able to sell your IPO stock if it is trading at a discount during pre-market trades.
To avoid these risks, consider investing in IPO stocks after the official IPO price has been announced and trading has begun. 

Questions: Can I invest in stocks without opening an account? Answer: No, you"
\end{tcolorbox}
\noindent\rule{\linewidth}{0.5pt}

\end{document}